\documentclass{article}
\usepackage{ijcai21}

\usepackage{times}
\usepackage{soul}
\usepackage{url}
\usepackage[hidelinks]{hyperref}
\usepackage[utf8]{inputenc}
\usepackage[small]{caption}
\usepackage{graphicx}
\usepackage{amsmath}
\usepackage{amsthm}
\usepackage{booktabs}
\usepackage{algorithm}
\usepackage{algorithmic}

\usepackage{amsfonts,amssymb}
\usepackage{multirow}

\usepackage{float}
\usepackage{subcaption}
\title{BiSTF: Bilateral-Branch Self-Training Framework for Semi-Supervised Large-scale Fine-Grained Recognition}

\author{
Hao Chang$^1$\and
Guochen Xie$^1$\and
Jun Yu$^1$\footnote{Contact Author}\And
Qiang Ling$^1$
\\
\affiliations
$^1$University of Science and Technology of China\\
\emails
\{changhaoustc, xiegc\}@mail.ustc.edu.cn,
\{harryjun, qling\}@ustc.edu.cn
}

\begin{document}

\maketitle

\begin{abstract}

Semi-supervised Fine-Grained Recognition is a challenge task due to the difficulty of  data imbalance, high inter-class similarity and domain mismatch.  Recent years, this field has witnessed great progress and many methods has gained great performance. However, these methods can hardly generalize to the large-scale datasets, such as Semi-iNat, as they are prone to suffer from noise in unlabeled data and the incompetence for learning features from imbalanced fine-grained data. In this work, we propose Bilateral-Branch Self-Training Framework (BiSTF), a simple yet effective framework to improve existing semi-supervised learning methods on class-imbalanced and domain-shifted fine-grained data. By adjusting the update frequency through stochastic epoch update, BiSTF iteratively retrains a baseline SSL model with a labeled set expanded by selectively adding pseudo-labeled samples from an unlabeled set, where the distribution of pseudo-labeled samples are the same as the labeled data. We show that BiSTF outperforms the existing state-of-the-art SSL algorithm on Semi-iNat dataset.

\end{abstract}

\section{Introduction}

With the emergence of research on deep Convolutional Neural Networks (CNNs), the performance of image recognition has witnessed incredible progress. The performance benefit is mainly conferred by the use of the large datasets and therefore easily lead to significant cost, since labeling data often requires human labor.

\begin{figure}[htbp]
     \centering
     \includegraphics[scale=0.6]{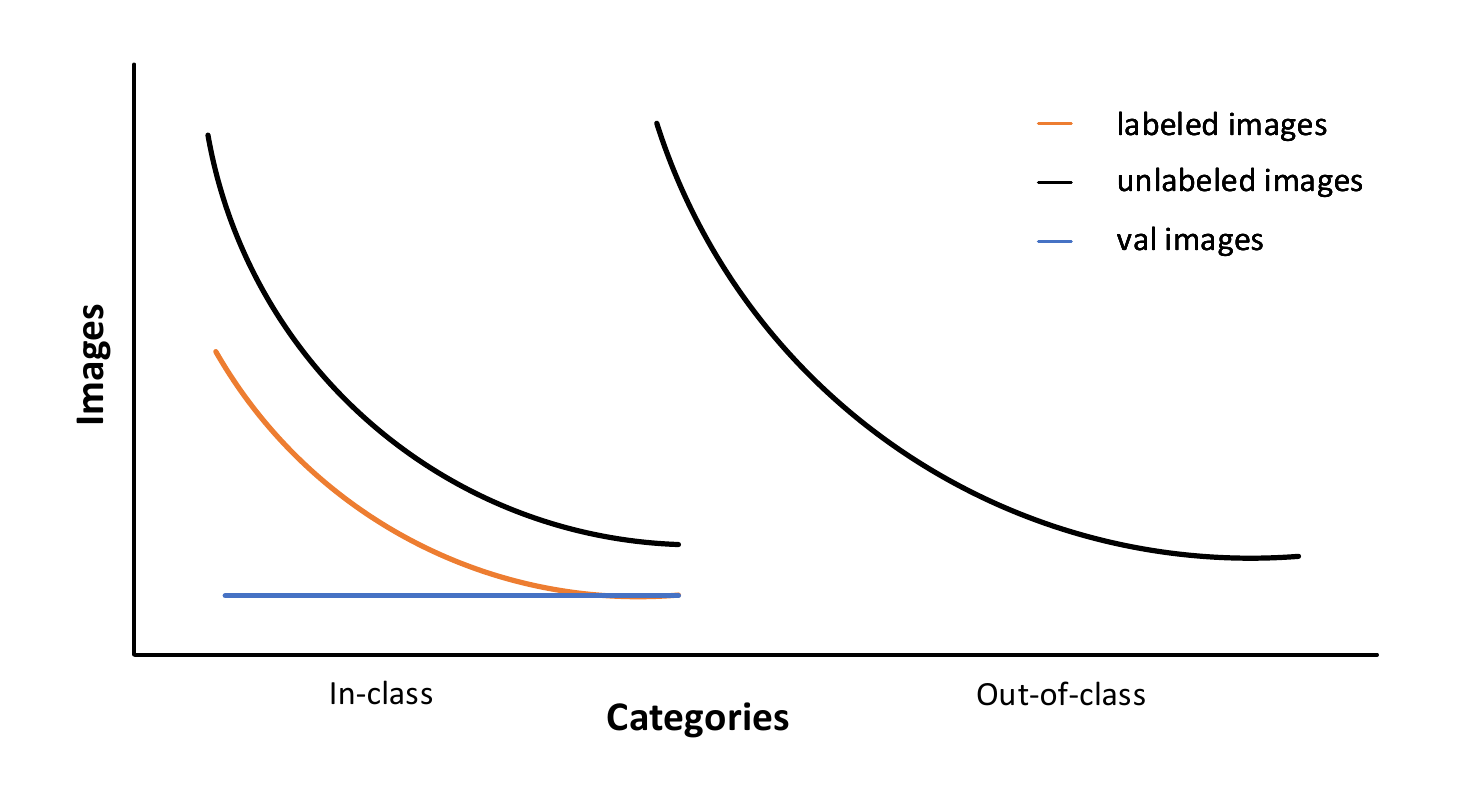}
     \caption{{\bf Class imbalance and domain mismatch on Semi-iNat Dataset.} Both labeled and unlabeled sets are class-imbalanced and domain-shifted, where the most majority class has 16× more samples than the most minority class. The validation and test set remains balanced.}
     \label{fig1}
\end{figure}

Semi-supervised learning (SSL) mitigates the requirement for labeled data with utilizing unlabeled data. A common assumption, which is often made implicitly during the construction of SSL benchmark datasets, is that the class distribution of labeled and unlabeled data with same categories are balanced. However, in many realistic scenarios, this assumption holds untrue, for example, that Semi-Supervised iNaturalist dataset(Semi-iNat)~\cite{su2021semi} has long-tailed data distribution, and more challengingly, it does not make a distinction between in-class and out-of-class unlabeled data.

Fine-grained image recognition and supervised learning on imbalanced data has been widely explored. The challenges of fine-grained recognition are mainly two-fold: discriminative region localization and fine-grained feature learning from those regions. Previous research has made impressive progresses by introducing part-based recognition frameworks, which relies on labels to identify possible object regions and extract discriminative features from each region. It is commonly observed that models trained on imbalanced data are biased towards majority classes which have numerous examples, and away from minority classes which have few examples. Various solutions have been proposed to help alleviate bias, such as re-sampling~\cite{buda2018systematic,byrd2019effect} and re-weighting~\cite{cui2019class,cao2019learning}. All these methods rely on labels to re-balance the biased model without domain mismatch between the labeled and unlabeled data.

In contrast, semi-supervised fine-grained image recognition on class-imbalanced and domain-mismatched data has been understudied. In fact, class imbalance and domain mismatch pose further challenges in SSL where missing label information precludes rebalancing the unlabeled set and distinguish between in-class and out-of-class unlabeled data. Pseudo-labels for unlabeled data generated by a model trained on labeled data are commonly leveraged in SSL algorithms. However, pseudo-labels can be problematic if they are generated by an initial model trained on imbalanced and domain-shifted data, as well as biased toward majority classes and out-of-class data: in addition to increasing noise due to out-of-class unlabeled data mistaken for in-class categories, subsequent training with such biased pseudo-labels intensifies the bias and deteriorates the model quality. The majority of existing SSL algorithms have not been thoroughly evaluated on class-imbalanced and domain-mismatched data. Besides, these algorithms all research on standard SSL image recognition benchmarks~\cite{tarvainen2017mean,berthelot2019mixmatch,sohn2020fixmatch,xie2020self} instead of fine-grained image recognition benchmarks.

In this work, we investigate SSL in the context of domain-shifted and imbalanced semi-supervised fine-grained recognition, as illustrated in Fig.\ref{fig1}. We observe that the undesired performance of existing SSL algorithms on Semi-iNat dataset is mainly due to the highly similar classes, class imbalance, and domain mismatch between the labeled and unlabeled data. We calculate the results on Semi-iNat dataset produced by FixMatch~\cite{sohn2020fixmatch} that is a representative SSL algorithm with state-of-the-art performance on balanced SSL benchmarks. In addition to obtaining low accuracy overall on the balanced test set, the model lacks the ability to identify fine-grained features and introduces noisy data.

With this in mind, this paper introduces a bilateral-branch self-training framework (BiSTF), which trains unbalanced data through a bilateral -branch structure, and samples pseudo-labeled data while maintaining the same data distribution through a stochastic epoch update strategy.  In order to improve the fine-grained learning ability of the model, BiSTF utilizes a backbone with an attention mechanism. Rather than updating the labeled set in each generation, we instead use a stochastic update epoch strategy to moderate the noise from out-of-class data in which the frequency of update increases as training progresses. In addition, to avoid the model being biased towards the majority class, the proposed method samples the unlabeled data with the same distribution as the labeled data set and add it into the labeled set to retrain an SSL model for next generation.

We show in experiments that BiSTF improves over baseline SSL method by a large margin. On Semi-iNat dataset~\cite{su2021semi}, our method outperforms FixMatch~\cite{sohn2020fixmatch} by as much as 10.25\% in accuracy. Extensive ablation study further demonstrates that our method particularly helps to improve ability to extract pseudo-labeled data from domain-shifted unlabeled data, making it a viable solution for class-imbalanced and domain-shifted semi-supervised fine-grained image recognition.

\section{Related work}

\subsection{Semi-supervised learning}

Recent years have observed a significant advancement of semi-supervised research~\cite{lee2013pseudo,miyato2018virtual,berthelot2019mixmatch,sohn2020fixmatch,xie2020self,xie2019unsupervised,laine2016temporal}. Many of these SSL methods share similar basic techniques, such as pseudo-labeling, or consistency regularization, combined with deep learning.  Pseudo-labeled~\cite{lee2013pseudo,sohn2020fixmatch} uses the pseudo-labeled target predicted by the model itself to train a classifier with unlabeled data. Consistency regularization~\cite{miyato2018virtual,laine2016temporal} promotes the consistency of predictions between different views through soft~\cite{miyato2018virtual,berthelot2019mixmatch,laine2016temporal} or hard~\cite{sohn2020fixmatch} pseudo-labels, thereby learning classifiers. The performance of recent SSL methods depends on the quality of pseudo-labels. However, none of the above works has studied SSL in an class-imbalanced dataset, in which model bias significantly threaten the quality of pseudo-labels.

\subsection{Class-imbalanced supervised learning}

Research on class-imbalanced learning has attracted increasing attention for supervised situation. Prominent work includes re-sampling~\cite{chawla2002smote,buda2018systematic} and re-weighting~\cite{khan2017cost,cui2019class}, which adjust the network training by rebalancing the contribution of each class in expectation closer to the test distributions, while others focus on re-weighting each instance~\cite{lin2017focal}. These methods assume that all labels of data fed into the model are available during training phase, and due to the missing label information in the SSL scenario, the performance is largely unknown.

\subsection{Class-imbalanced semi-supervised learning}

Although SSL has been extensively studied, it is still underexplored for class imbalanced data. Recently, ~\cite{yang2020rethinking} shown that the use of SSL and self-supervised learning can be beneficial to class imbalanced learning. ~\cite{hyun2020class} proposed a method to suppress the loss of minority classes by suppressing the consistency loss. Although these works have done some research on SSL under unbalanced data distribution, there is no more discussion for domain mismatch, neither for fine-grained recognition.

\section{Approach}

In this section, we first introduce Semi-iNat, a challenging dataset for semi-supervised recognition. Next, we set up the problem and introduce the baseline supervised algorithm   leveraged by the official and SSL algorithms. Then we investigate the misclassified and biased behavior of existing SSL algorithms on Semi-iNat. Based on these observations, we propose a bilateral-branch self-training framework that fine-tunes model by in-class supervised branch to avoid noise from mistaken pseudo-labels, and takes advantage of, rather than suffers from, the model’s bias to enhance performance on minority classes.

\subsection{Semi-iNat}
\label{section31}

Different from standard SSL image recognition benchmarks, Semi-iNat dataset~\cite{su2021semi} is full of challenges for semi-supervised recognition with fine-grained categories, a long-tailed distribution of classes, and domain mismatch between labeled and unlabeled data.

{\bf Classes} Semi-iNat contains images of species from three kingdoms in the natural taxonomy: Animal, Plants, and Fungi (Tab.\ref{tab1}).

{\bf Split} This dataset is at a larger scale for a total of $\approx330$k images. Specially, it is split into two sets $C_{in}$ with 810 in-class species and $C_{out}$ with 1629 out-of-class species. For each species in $C_{in}$, $5/10/10$ images are selected for validation, public test, and private test set. Among the rest of the images, around $90\%$ of the images are sampled as unlabeled data $U_{in}$ and the rest as labeled data $L_{in}$. In addition, each class is guaranteed to have at least 5 labeled images. For species in $C_{out}$, all of them are included in $U_{out}$. The two sets of unlabeled data are then combined $U = U_{in} \bigcup U_{out}$, and more challengingly, no domain labels are provided but coarse taxonomic labels for the unlabeled data are provided, such as kingdom and phylum. The statistics of the class distribution has shown in Fig.\ref{fig1}.

\begin{table}
\centering
\begin{tabular}{ccrr}
\toprule
    {\bf Kingdom}    & {\bf Phylum}  & $C_{in}$  & $C_{out}$ \\
\midrule
Animalia       & Mollusca  &11  &24 \\
       & Chordata &  113  &    228  \\
      & Arthropoda  &  301 &  605   \\
          & Echinodermata  & 4 &  8   \\

Plantae         &  Tracheophyta  &  336 &  674   \\
         & Bryophyta  &  6 &   12   \\

Fungi         & Basidiomycota  & 29 &   58   \\
              & Ascomycota  & 10 &   20   \\

\bottomrule
\end{tabular}
\caption{{\bf The number of species in the taxonomy.} For each phylum, around one-third of the species are selected for the in-class set $C_{in}$ and the rest for the out-of-class set $C_{out}$.}
\label{tab1}
\end{table}

\subsection{Problem setup and baselines}

First of all, we set up the problem of domain-shifted and class-imbalanced semi-supervised fine-grained recognition.

Considering the labeled set $L_{in} = \{(x_{ln}, y_{ln}): n \in (1, \cdots , N)\}$, let $x_{ln}$ denote a training sample and $y_{ln} \in {1, 2,\cdots,C}$ is its corresponding label for a C-class recognition task. Without loss of generality, this paper assumes that the number of training examples in $L_{in}$ of class c is denoted as $N_{c}$, i.e., $\sum_{c=1}^{C}N_{c}=N$, and sorted by cardinality in descending order, i.e., $N_{1} \geq N_{2} \geq \cdots \geq N_{C}$. Evidently, due to the long-tailed distribution, the marginal class distribution of $L_{in}$ is skewed, i.e., $N_{1} \gg N_{C}$. We use imbalance ratio to measure the degree of class imbalance, $\gamma=\frac{N_{1}}{N_{C}}$. Besides the labeled set $L_{in}$ , an unlabeled set $U = u_{m} : m \in (1, \cdots , M)$ that misses label information and does not match the domain with labeled set $L_{in}$ is also provided. Given sets $L_{in}$ and $U$, our goal is to learn a classifier $\mathbf{f} : x \rightarrow {1, \cdots , C}$ that generalizes well under the class-balanced public test and private test criterion.

The official of Semi-iNat presents a result of fully-supervised model on the labeled set $L_{in}$ using ResNet-50~\cite{he2016deep} models trained from ImageNet pre-trained model. They built this general recognition network with basic training strategies. Many existing state-of-the-art SSL methods assign a pseudo-label with the classifier’s prediction $\hat{y}_m = f(u_{m})$ to leverage unlabeled data.  The classifier is then optimized on both labeled and selected unlabeled samples with corresponding high-confidence pseudo-labels. Therefore, the quality of pseudo-labels is crucial to the performance of final SSL algorithms. These SSL algorithms work successfully on standard class-balanced benchmarks because the quality of the classifier improves during the training process, owing to the addition of pseudo-labeled data.

However, when the classifier is biased due to the shifted domain and a skewed class distribution at the beginning of training, the online pseudo-labels of unlabeled data may be even more biased, which may further aggravate the domain mismatch and class imbalance issue and result in weakened performance on labeled set.

\subsection{How baseline performs on Semi-iNat?}
\label{section33}

Instead of extending the protocol, that is utilizing various class-imbalanced ratios to produce long-tailed versions of benchmark datasets, such as CIFAR~\cite{krizhevsky2009learning}, and retaining a fraction of training data as labeled and the rest as unlabeled, Semi-iNat has been split into train, validation, public test, and private test set. We test FixMatch~\cite{sohn2020fixmatch}, one of the state-of-the-art SSL algorithms, which is designed for class-balanced data. By calculating validation recall, validation precision of each class and test precision on Semi-iNat dataset, we find that FixMatch only has limited help for fine-grained image recognition on Semi-iNat.

\begin{figure}[htbp]
     \centering
     \includegraphics[scale=0.6]{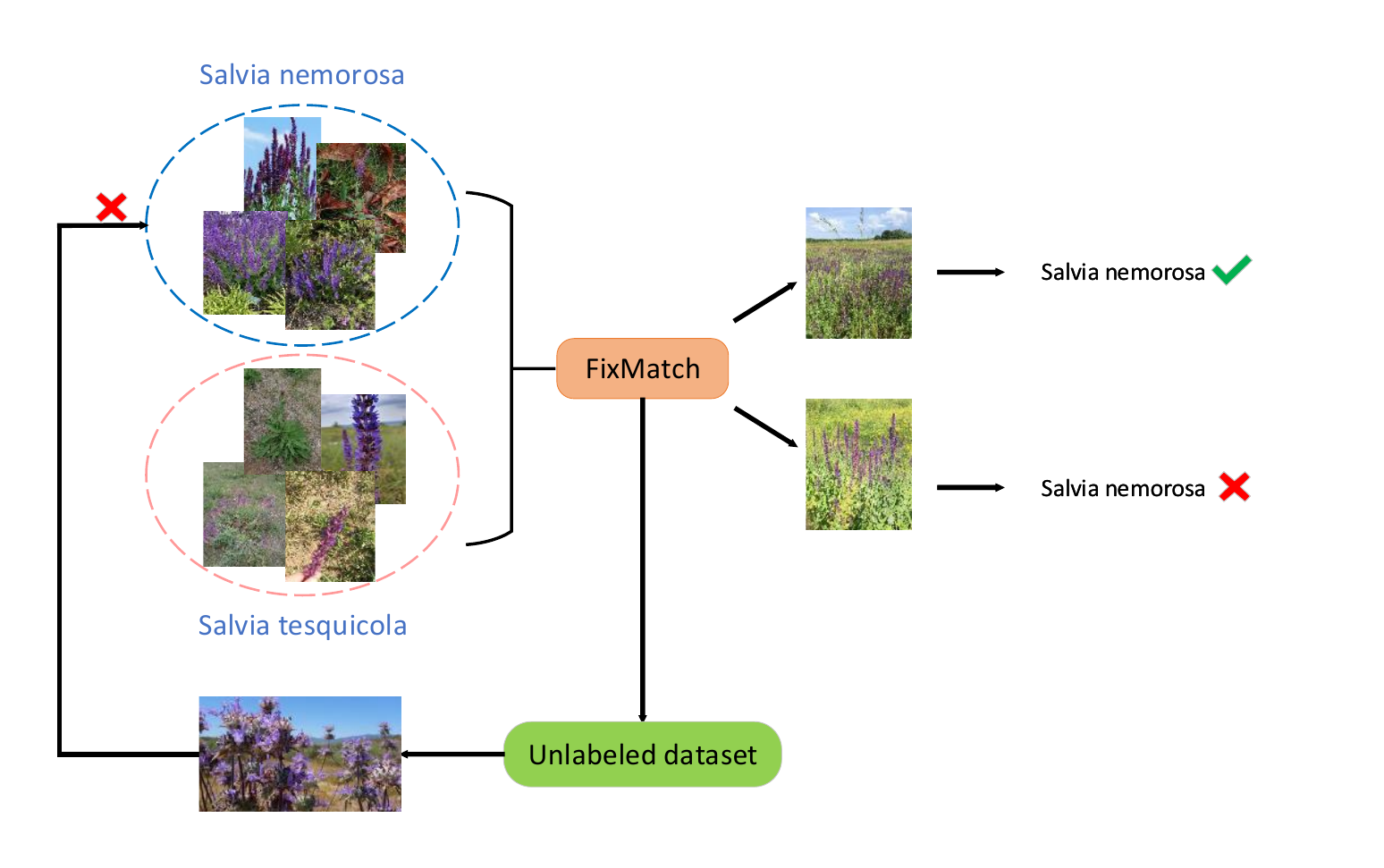}
     \caption{{\bf Performance of FixMatch on Semi-iNat.} An image of Salvia tesquicola is classified incorrectly due to highly similar classes. Besides, FixMatch introduces noise from out-of-class data during training phase.}
     \label{fig2}
\end{figure}

As shown in the Fig.\ref{fig2}, Salvia nemorosa speices and Salvia tesquicola speices are members of Salvia genus. However, during training phase, an image belonging to Salvia tesquicola is classified incorrectly as Salvia nemorosa by FixMatch, which shows that FixMatch has a weak learning ability for fine-grained features. When selecting pseudo-labeles from unlabeled data, we found that FixMatch introduces noise from out-of-class data, so that in the next iteration, the model is heavily biased towards out-of-class data. In addition, the quality of pseudo-labels is reduced due to class imbalance, also resulting in the poor performance on Semi-iNat. These empirical findings motivate us to improve the model's ability to learn fine-grained features and alleviate the impact of class imbalance and domain mismatch.

\begin{figure*}
     \centering
     \includegraphics[scale=0.5]{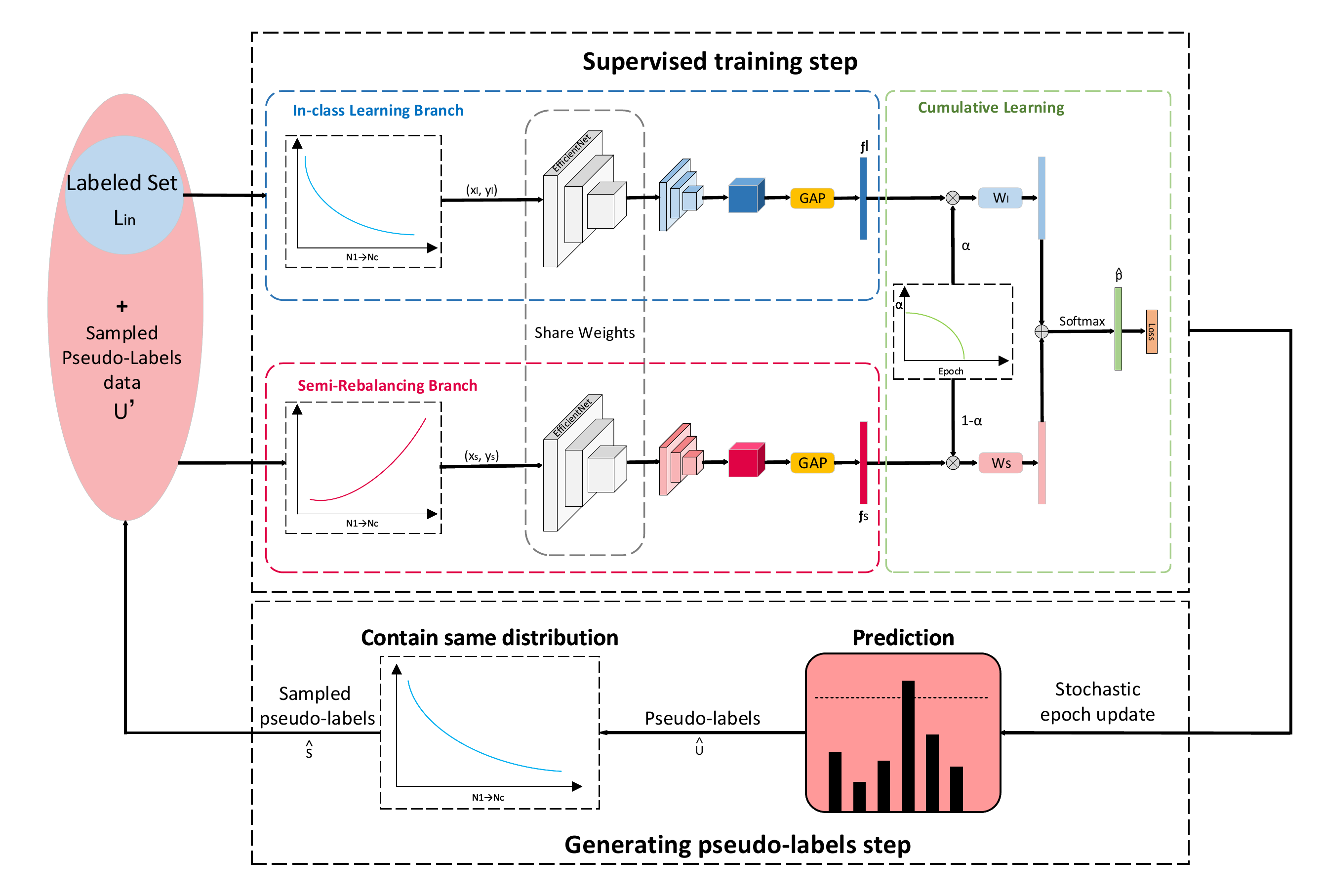}
     \caption{{\bf BiSTF(Bilateral-Branch Self-Training Framework).} By adjusting the update frequency through stochastic epoch update, BiSTF iteratively retrains a baseline SSL model with a labeled set expanded by adding pseudo-labeled samples from an unlabeled set, where pseudo-labeled samples contain same data distribution with the labeled dataset. See text for details.}
     \label{fig3}
\end{figure*}

To achieve this goal, we introduce BiSTF, a bilateral-branch self-training framework for domain-shifted and class-imbalanced semi-supervised fine-grained recognition illustrated in Fig.\ref{fig3}.

\subsection{Bilateral-branch self-training framework}

In SSL, self-training as an iterative method is widely used. It trains the model for multiple generations, in which each generation involves two training steps, supervised training step and generating pseudo-labels step.

As shown in Fig.\ref{fig3}, our BiSTF consists of abovementioned two main steps. In supervised training step, the model contains three main components. Concretely, we design two branches for in-class representation learning and semi-supervised classifier learning, termed “in-class learning branch” and “semi-rebalancing branch”, respectively.

Existing SSL algorithms usually ignore the subtle but discriminative features in fine-grained recognition, hence a network structure with an attention mechanism is necessary. In order to satisfy the needs of fine-grained recognition, both branches use the same EfficientNet~\cite{tan2019efficientnet} network structure, where SENet~\cite{hu2018squeeze} can effectively capture the fine-grained features rather than residual network structure, and share all the weights except for the last MBConv block.

For the bilateral branches, we separately apply uniform and reversed samplers proposed in BBN~\cite{zhou2020bbn} to each of them and obtain two samples $(x_{l}, y_{l})$ and $(x_{s}, y_{s})$ as the input data, where $(x_{l}, y_{l})$ is from labeled set $L_{in}$ for the in-class learning branch and $(x_{s}, y_{s})$ is from the union set $U'$ of $L_{in}$ and sampled pseudo-labels $\hat{S}$ generated by next step for the semi-rebalancing branch. The uniform sampler retains the characteristics of original distributions, and therefore benefits the representation learning. While, the reversed sampler aims to alleviate the extreme imbalance and particularly improve the recognition  accuracy on minority class. Crucially, the two samplers with different sampling methods are the biggest difference between the two branches. Then, two samples are fed into corresponding branch, and by global average pooling the feature vectors  $ \mathbf{f}_{l} \in \mathbb{R}^D$ and $ \mathbf{f}_{s} \in \mathbb{R}^D$ can be acquired.

Furthermore, we also introduce the specific cumulative learning strategy in to shift the mode’s learning “attention” in the supervised training step. Different from BBN, the adaptive trade-off parameter $\alpha$ directly affects the model's ability to bias towards in-class data and re-balance the data by controlling the weights for $\mathbf{f}_{l}$ and $\mathbf{f}_{s}$. The outputs will be integrated together by element-wise addition after sending the weighted feature vectors $\alpha \mathbf{f}_{l}$ and $(1-\alpha)\mathbf{f}_{s}$ into the classifiers $\mathbf{W}_{l}\in \mathbb{R}^{D\times C}$ and $\mathbf{W}_{s}\in \mathbb{R}^{D\times C}$ respectively. The output logits are formulated as

\begin{align}
    z=\alpha \mathbf{W}_{l}^ \top \mathbf{f}_{l}+(1-\alpha)\mathbf{W}_{s}^ \top \mathbf{f}_{s}
\end{align}%

where $z\in\mathbb{R}^C$ is the predicted output, i.e., $[z_{1}, z_{2},\cdots,z_{C} ]^\top $. For each class $i \in \{1, 2,\cdots,C\}$, the the probability of the class  is calculated by softmax function

\begin{align}
    \hat{\mathbf{p}_{i}}=\frac{e^{z_{i}}}{\sum_{j=1}^{C}e^{z_{i}}}
\end{align}%

Then, we generally denote the output probability distribution as $\hat{\mathbf{p}} =[\hat{p_{1}}, \hat{p_{2}},\cdots, \hat{p_{C}} ]^\top$, $E(\cdot,\cdot)$ as the cross-entropy loss function. Thus, our model generates a weighted cross-entropy recognition loss, which is illustrated as

\begin{align}
    L=\alpha E(\mathbf{\hat{\mathbf{p}}},y_{l})+(1-\alpha)E(\mathbf{\hat{\mathbf{p}}},y_{s})
\end{align}%

As observed in Sec.\ref{section33}, out-of-class unlabeled data will easily be introduced in the early stage of training and cause interference to the model. Therefore, instead of iteration update strategy of FixMatch, we propose “stochastic epoch update” strategy when supervised training step has thoroughly trained. Specifically, at the beginning of training, to avoid introducing out-of-class noise to affect the model performance, we perform pseudo-labeling on the unlabeled dataset with a small probability and selectively add them to the training phase, and then the probability of epoch updating gradually increases. We define whether to update or not as a flag $F_{update}$.

Pseudo-labeling leverages the idea of using the model obtained from the first step to generate artificial labels for unlabeled data. Specifically, when the largest class probability fall above a predefined threshold, this hard label will be retained as pseudo-label. Letting $q_{m}=p_{m}(y|u_{m})$, pseudo-labeling uses the following  function:

\begin{align}
    \hat{y}_{m}=f(max(q_{m})\geq \tau)
\end{align}%

where $\tau$ is the threshold. The pseudo-labeled set $\hat{U}=\{(u_{m},\hat{y}_{m})\}_{m=1}^M$.

To accommodate the class-imbalance, this paper proposes “contain same distribution” strategy, that instead expands the labeled set with a selected subset $\hat{S}\subset \hat{U}$, i.e., $U'=L_{in}\cup \hat{S}$, rather than with all samples in $\hat{U}$. The biased pseudo-labels generated by an initial model trained on imbalanced data will intensify the bias. Consequently, in the selection process, we follow the strategy of keeping the selected pseudo-labeled data distribution consistent with the in-class data distribution, which avoids that the data distribution of $U'$ is gradually biased towards majority classes. For the next generation, the labeled set $L_{in}$ and the union set $U'$ will be fed into the “in-class learning branch” and “semi-rebalancing branch”, respectively.

\section{Experiments}

\subsection{Datasets and empirical settings}

{\bf Semi-iNat}  In addition to the introduction of Semi-iNat in Sec.\ref{section31}, a parameter that is critical to our experiments is the maximum imbalance rate with the value of $\gamma=16$. In this paper, the official splits of train, validation, public test, and private test set are utilized for fair comparisons.

\subsection{Implementation details}

To be fair, we train the ResNet-50~\cite{he2016deep} leveraged by the official as our backbone network by standard mini-batch stochastic gradient descent (SGD) with momentum of 0.9, weight decay of $1\times10^{-4}$. Our experiments follow the data augmentation strategies proposed in FixMatch~\cite{sohn2020fixmatch}: resize image to $224\times224$, random horizontal flip with $50\%$ probability, randomly resizecrop a $224\times224$ patch from the original image or its horizontal flip with  scale from 0.2 to 1.0 and ratio from 0.75 to 1.33, as well as RandAugment~\cite{cubuk2020randaugment} keeping the same settings in FixMatch. We train all the models on a single NVIDIA A100 GPU with batch size of 64. The initial learning rate is set to 0.01 and the learning rate during subsequent training is decayed by ReduceLROnPlateau scheduler with patience of 5.

\subsection{Main results}

First, we compare our model with baseline reproduced according to the official and FixMatch, and present the results in Tab.\ref{tab1}. Due to the utilization of data augmentation, the accuracy of the baseline reproduced by us is improved by 6.21$\%$ over the result by the official that is for reference only. Although FixMatch performs reasonably well on public and private test set, its improvement is not as obvious as in the basic SSL benchmark. In contrast, BiSTF improves the accuracy of FixMatch and achieves as much as 1.79$\%$ absolute performance gain.

\begin{table}
\centering
\begin{tabular}{lccc}
\toprule
        & Val  & Public test  & Private test \\
\midrule
Official       & $31.00\%$  &--  &-- \\
Baseline       & $37.21\%$ & $34.86\%$  &   $35.60\%$  \\
Fixmatch       & $37.53\%$  & $36.15\%$ &  $36.40\%$   \\
BiSTF          & $\mathbf{39.31\%}$  & $\mathbf{37.90\%}$ & $\mathbf{38.19\%}$   \\
\bottomrule
\end{tabular}
\caption{{\bf We compare BiSTF with baseline methods.}  All models are trained with the same settings for fair comparison.}
\label{tab1}
\end{table}

We also observe that our model works particularly well and achieves 1.75$\%$ and 1.79$\%$ accuracy gain on public test and private test data, respectively. We hypothesize the reason is that by stochastic epoch update strategy our model finds more correctly pseudo-labeled samples to augment the labeled set instead of iteration update strategy of FixMatch. In addition, by observing the performance on the validation set, it shows that the ability of BiSTF to learn fine-grained features has been improved.

We further report the performance of BiSTF with different backbones and image sizes in Tab.\ref{tab2}. After resizing images to $600\times600$, this paper first directly evaluates several common backbones, including Resnet101~\cite{he2016deep}, ResneXt101~\cite{xie2017aggregated}, EfficientNet-b5-7~\cite{tan2019efficientnet}. All the backbones are able to further boost the performance by another few points, resulting in 5.0 to 10.25$\%$ absolute accuracy improvement compared to BiSTF with backbone of Resnet50. Introducing the noisy student~\cite{xie2020self} to EfficientNet, the results of BiSTF can be further improved. Among these backbones, EfficientNetb7$\_$ns achieves the best performance, so we take it as the final baseline. Finally, applying 5-fold cross-validation after expanding the validation data to training data, our BiSTF model further gives accuracy gains, producing the best results.

\begin{table}
\centering
\begin{tabular}{lccc}
\toprule
        & Val  & Public test  & Private test \\
\midrule
ResNet50                  & $44.33\%$  & $42.18\%$ &$42.85\%$ \\
ResNet101                  & $49.42\%$  & $47.03\%$ &$47.52\%$ \\
ResNeXt101                  & $61.86\%$  & $59.96\%$ &$60.12\%$ \\
EfficientNet-b5                  & $53.78\%$  & $52.33\%$ &$53.10\%$ \\
EfficientNet-b7               & $64.77\%$ & $62.63\%$  &   $63.47\%$  \\
EfficientNet-b7\_ns  & {\multirow{2}{*}{--}}  & {\multirow{2}{*}{$\mathbf{77.00\%}$}} & {\multirow{2}{*}{$\mathbf{77.48\%}$}}    \\
(expand-data)\\
\bottomrule
\end{tabular}
\caption{{\bf Top-1 Accuracy (\%) of BiSTF with different backbones on Semi-iNat.}}
\label{tab2}
\end{table}

\subsection{Ablation studies}

We perform an extensive ablation study to evaluate and understand the contribution of critical component in BiSTF. The experiments in this section are all performed with BiSTF on Semi-iNat.

\begin{figure}[htbp]
	\centering
	\begin{subfigure}{0.45\linewidth}
		\centering
		\includegraphics[width=1.1\linewidth]{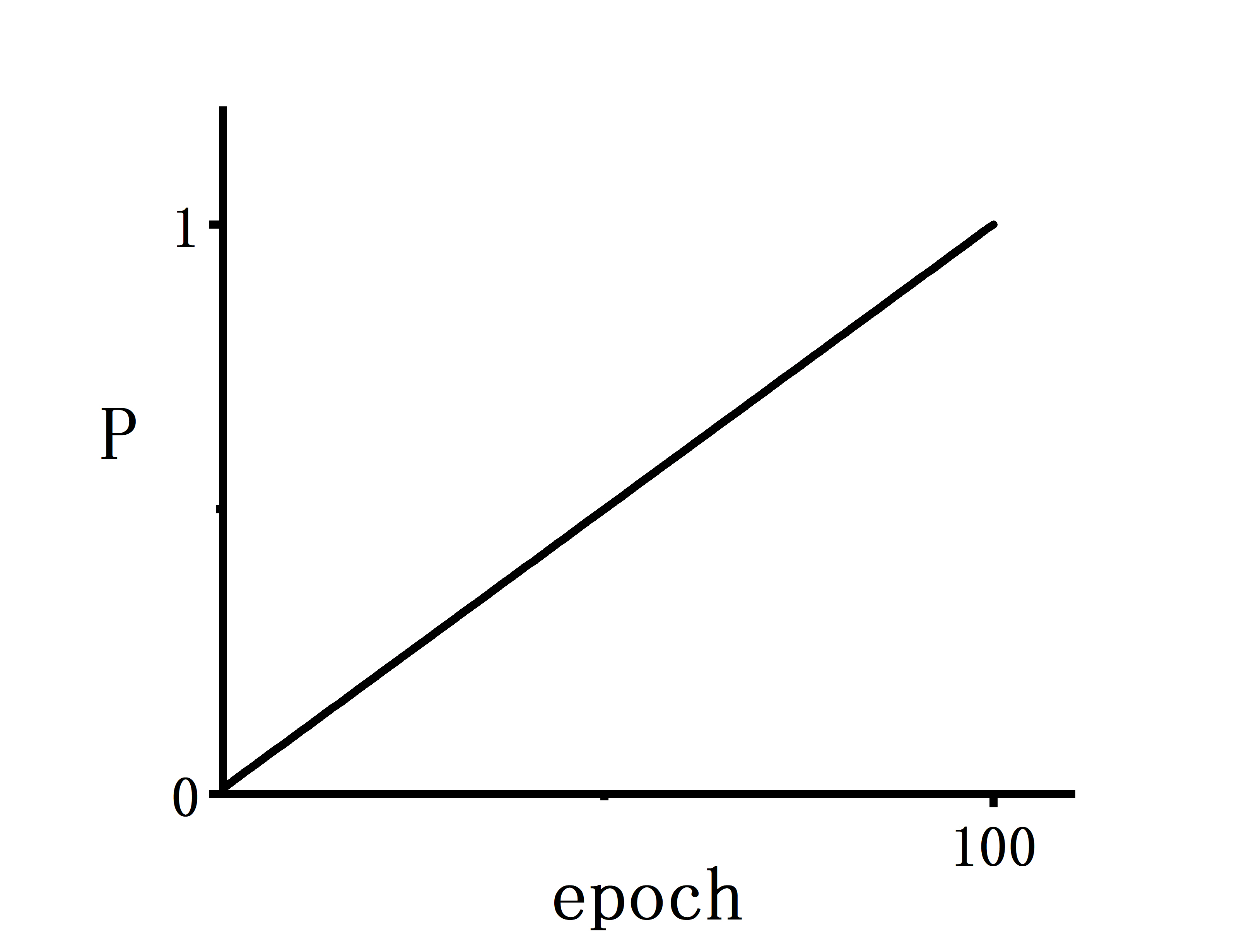}
		\caption{Linear}
		\label{chutian3}
	\end{subfigure}
	\centering
	\begin{subfigure}{0.45\linewidth}
		\centering
		\includegraphics[width=1.1\linewidth]{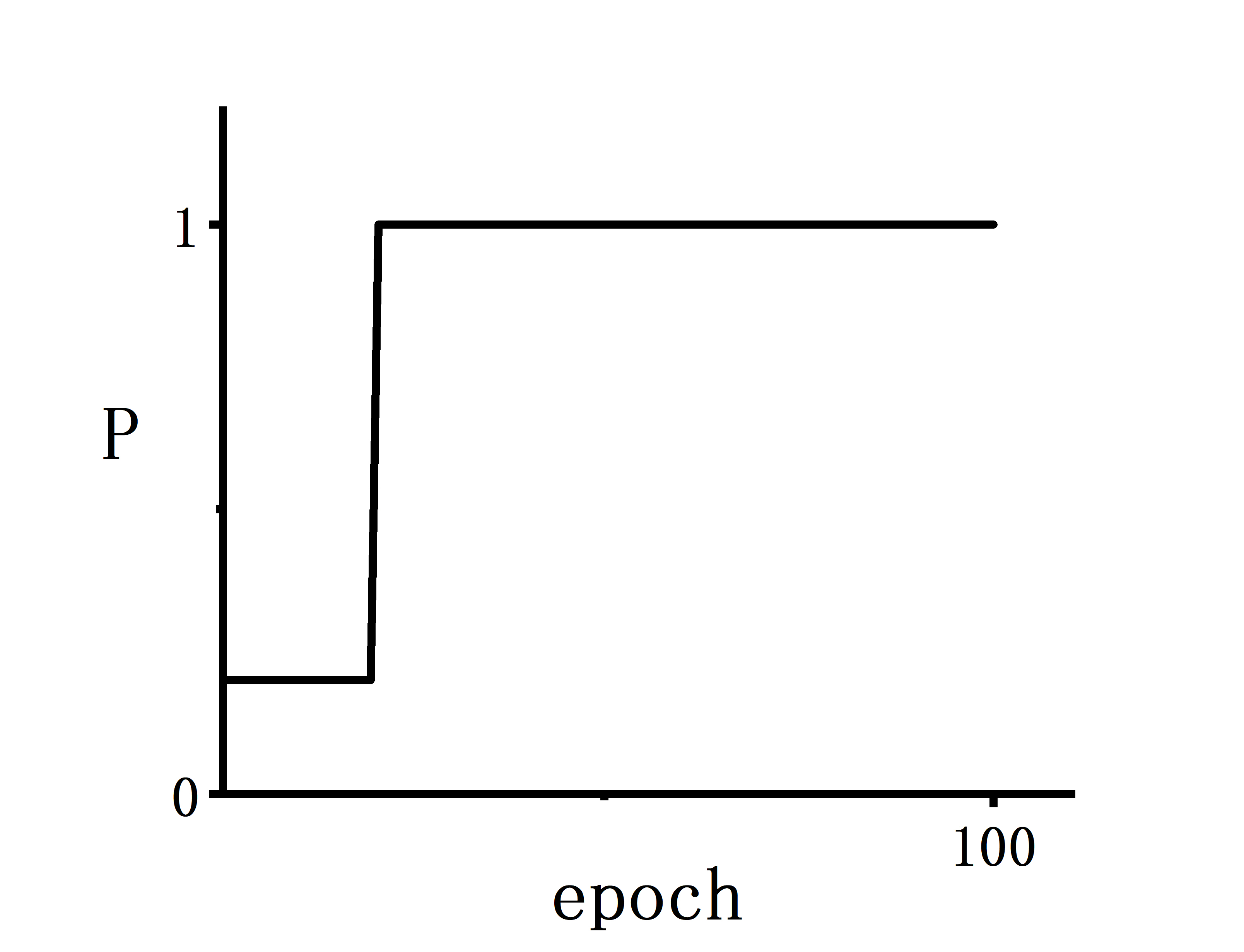}
		\caption{Seperated stage}
		\label{chutian3}
	\end{subfigure}
	\caption{{\bf Stochastic epoch update strategy.} Update probability is a)linear or b)seperated stage with epoch}
	\label{fig4}
\end{figure}

{\bf Effect of update probability}. BiSTF introduces the "Stochastic epoch update strategy" that controls the update frequency. Fig.\ref{fig4} shows how update strategy influences performance over generations.

When P = 1 in the whole process of training, our method updates the dataset in each generation. Besides, two other  update strategies are also tested. Specifically, the update probability varies linearly and in separated stages with the epoch, respectively. To show the source of accuracy improvements, in Tab.\ref{tab3} we present accuracy on the validation set of Semi-iNat. The results suggest that various stochastic update strategies affect the learning ability of the model to a certain extent by changing the pseudo-labeled data sampled for training, where the linear strategy produces the best result.

\begin{table}
\centering
\begin{tabular}{lccc}
\toprule
Stochastic epoch update strategy     & Val  \\
\midrule
All                  & $38.00\%$  \\
Seperated stage      & $38.59\%$  \\
Linear               & $\mathbf{39.31\%}$  \\

\bottomrule
\end{tabular}
\caption{{\bf Top-1 Accuracy (\%) of BiSTF with different “Stochastic epoch update” strategies on Semi-iNat.}}
\label{tab3}
\end{table}

\section{Conclusion}

In this work, we present a bilateral-branch self-training framework, named BiSTF for domain-shifted and imbalanced semi-supervised fine-grained recognition. BiSTF is motivated by the observation that in addition to ignoring the subtle but discriminative features, existing SSL algorithms are vulnerable to class imbalance and domain mismatch. BiSTF iteratively refines a baseline SSL model with a labeled set expanded by adding pseudo-labeled samples from an unlabeled set, where pseudo-labeled samples contain same data distribution with the labeled dataset. Over generations of self-training, the model becomes less biased towards majority classes and out-of-class data, focusing more on in-class data. Extensive experiments on Semi-iNat datasets demonstrate that the proposed BiSTF outperform the existing state-of-the-art SSL algorithm.

\bibliographystyle{named}
\bibliography{ijcai21}

\end{document}